\documentclass[runningheads]{llncs}

\usepackage{eccvabbrv}

\usepackage{graphicx}
\usepackage{booktabs}

\usepackage[accsupp]{axessibility}  

\usepackage{orcidlink}
\usepackage{wrapfig}
\usepackage{mathtools}

\begin{document}

\title{Personalized Video Relighting With an At-Home Light Stage} 


\author{Jun Myeong Choi\orcidlink{0009-0005-3185-1400} \and
Max Christman\orcidlink{0000-0002-8090-2797} \and
Roni Sengupta\orcidlink{0000-0001-5914-3469}}

\authorrunning{J. Choi et al.}

\institute{University of North Carolina at Chapel Hill, NC 27514, USA\\
\email{\{chedgekr,mrlc,ronisen\}@cs.unc.edu}}

\maketitle


\begin{abstract}
  In this paper, we develop a personalized video relighting algorithm that produces high-quality and temporally consistent relit videos under any pose, expression, and lighting condition in real-time. Existing relighting algorithms typically rely either on publicly available synthetic data, which yields poor relighting results, or on actual light stage data which is difficult to acquire. We show that by just capturing recordings of a user watching YouTube videos on a monitor we can train a personalized algorithm capable of performing high-quality relighting under any condition. Our key contribution is a novel image-based neural relighting architecture that effectively separates the intrinsic appearance features - the geometry and reflectance of the face - from the source lighting and then combines them with the target lighting to generate a relit image. This neural architecture enables smoothing of intrinsic appearance features leading to temporally stable video relighting. Both qualitative and quantitative evaluations show that our architecture improves portrait image relighting quality and temporal consistency over state-of-the-art approaches on both casually captured `Light Stage at Your Desk' (LSYD) and light-stage-captured `One Light At a Time' (OLAT) datasets. Source code is available at \url{https://github.com/chedgekorea/relighting}
\end{abstract}

\section{Introduction}
\label{sec:intro}

With the recent rise in popularity of video conferencing for business, educational, and personal activities, there is a significant demand for improving facial lighting. Virtually relighting our images and videos helps us improve the appearance of our faces without requiring explicit studio-quality lighting in a dedicated space or any specialized lighting expertise.  Recent advances in deep neural networks have renewed interest in the problem of virtual relighting.

\begin{figure*}[!h]
  \centering

  \includegraphics[width=\linewidth]{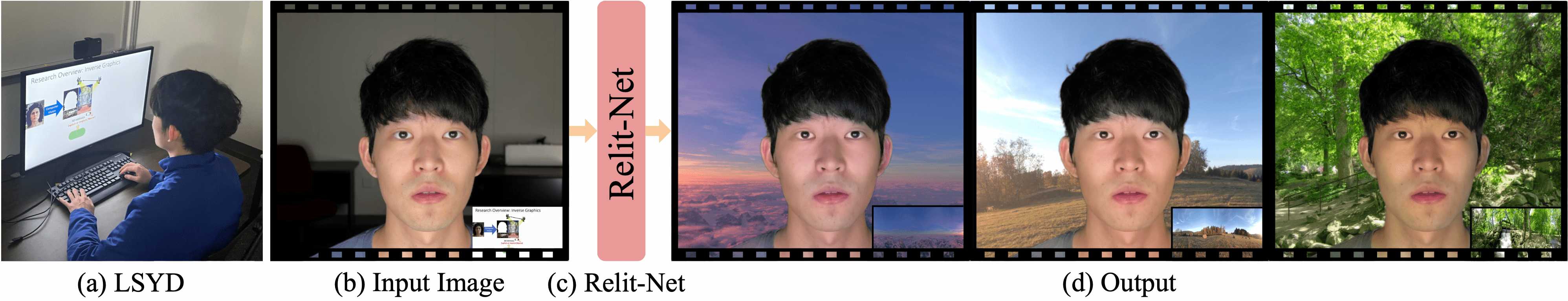}
  \caption{We learn a personalized relighting algorithm that generates temporally consistent and high-quality portrait videos under different lighting. We train the network using recordings of users watching YouTube on a monitor, thereby creating a Light Stage at Your Desk (LSYD). We project a portion of the LDR environment map with a 180 $^{\circ}$ FoV as the monitor light, while a portion of the remaining 90 $^{\circ}$ FoV is mapped as the background. We can achieve a harmonization effect with the virtual background. }
  \label{fig:teaser}

\end{figure*}

Training a deep neural network for relighting requires extensive training data that includes source images paired with relit target images. One way of acquiring this data is by using a large spherical rig with numerous lights and cameras, known as a light stage~\cite{Light_stage}. While light stage data has been shown to produce high-quality relighting results~\cite{Google,video_paper6,explicit_paper3,super_resolution_paper,explicit_paper1,shadow_manipulation}, the limited availability of datasets, trained models, and access to the light stage itself has impeded further research. For example, \textit{One Light At a Time} (Dynamic-OLAT)~\cite{video_paper6} is the only publicly available light stage relighting dataset consisting of four individuals only. As a result, researchers have often turned to synthetic data to train their relighting algorithms~\cite{sengupta2018sfsnet,ratio_paper1,ratio_paper2,body_relight}. Unfortunately, existing synthetic data compromises the quality of relit images.

We draw inspiration from recent work by Sengupta \etal ~\cite{Roni,my3dgen} and develop a personalized relighting model by capturing a single user's appearance while being lit by a computer monitor. However, the results and applications of Sengupta \etal ~\cite{Roni} have several limitations. First, it requires capturing users with fixed poses and expressions, an unrealistic requirement for actual users that impedes casual capture.  Second, the relighting algorithm requires a dark room with negligible ambient light, limiting the environment in which this can be applied. Third, the resulting relit video is temporally unstable, exhibiting significant flickering artifacts, making it unsuitable for relighting Zoom calls. Lastly, it requires  knowledge of source lighting and thus is unable to relight arbitrary portrait images captured in the wild.

In this paper, we show that casually captured light stage data is sufficient to develop a high-quality temporally consistent video portrait relighting algorithm that works under arbitrary conditions (i.e. pose, expression, and ambient lighting) in real-time ($\sim$45 fps). To that end, we create our own casually captured light stage dataset with varying pose, expression, and lighting, called \textit{Light Stage at Your Desk} (LSYD). Our key contribution is a neural image-based relighting architecture, based on the commonly used U-Net~\cite{Google,ratio_paper1,explicit_paper1,explicit_paper2,super_resolution_paper}, that better disentangles the source lighting from the user's intrinsic facial appearance (shape and reflectance) and then adds back the target lighting to generate a relit image. Existing image-based relighting architectures \cite{Google,Roni} fail to accurately separate source lighting information from intrinsic appearance features in the encoder, leading to inconsistent and temporally unstable video relighting. To this end, we introduce the \textit{light-conditioned feature normalization} (LCFN) module, which performs relighting and also predicts the source lighting from an input image. The LCFN module also enables temporal stability by performing exponential smoothing of \textit{de-lit} intrinsic appearance features and facilitates relighting of any arbitrary portrait image with unknown source lighting. We also improve the data pre-processing pipeline from Sengupta \etal \cite{Roni} to make the relighting algorithm more robust to pose, expression, and ambient lighting conditions.

We compare our relighting network with two other algorithms: Sun \etal \cite{Google}, which was originally trained on light stage data (OLAT), and Sengupta \etal \cite{Roni}, which was originally trained on casually captured data (albeit with fixed pose, expression, and no ambient lighting). For a fair comparison, we train all algorithms for personalized relighting using the same data pre-processing steps and loss functions on 5 individuals from our LSYD dataset and 4 individuals from OLAT \cite{video_paper6}. Our network outperforms Sun \etal \cite{Google} and Sengupta \etal \cite{Roni} by 22.3\% and 23.6\% respectively on the LSYD dataset and by 23.5\% and 25.6\% on the OLAT dataset, in terms of LPIPS. Qualitatively our method produces superior relighting in terms of color and quality. We further show that our approach is more temporally consistent, leading to less flickering than Sun \etal \cite{Google} or Sengupta \etal \cite{Roni}. Detailed ablation studies show that LCFN and source monitor prediction improves relighting quality, feature, and source monitor smoothing improves temporal consistency, and data pre-processing improves robustness to pose and expression.

\begin{figure*}[tb]
  \centering
  \includegraphics[width=\linewidth]{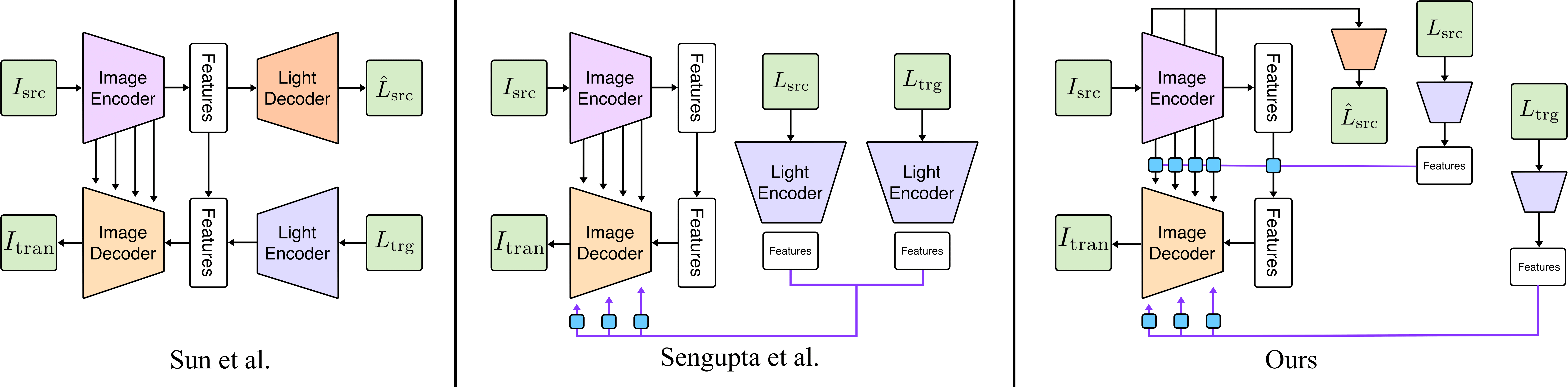}

  \caption{We highlight the key structural differences between our relighting architecture and that of \cite{Google,Roni}. Our approach removes source lighting information from input image features and only propagates intrinsic appearance (geometry and reflectance) features from the encoder to decoder, which results in better relighting quality and more temporal stability. In contrast \cite{Google,Roni} propagates entire image features from the encoder to the decoder without `de-light', and expects the decoder to remove source lighting and add target lighting information.}
  \label{fig:architecture_difference}

\end{figure*}

\noindent To summarize, our main contributions are as follows: 

$\bullet$ We show that casually captured \textit{Light Stage at Your Desk} (LSYD) data can be used to build a high-quality temporally consistent personalized video relighting algorithm without requiring access to an expensive light stage setup. 

$\bullet$ We introduce a novel video relighting architecture that separates the source lighting from the user's intrinsic appearance features and then adds back the target lighting, leading to improved relighting and temporal consistency for videos. 

$\bullet$ While our relighting network focuses on `at home' Light Stage (LSYD dataset) outperforming state-of-the-art algorithms, we also perform equally well on actual Light Stage captured OLAT \cite{video_paper6} datasets, and any arbitrary portrait image captured `in the wild'.

\section{Related Work}
Portrait relighting methods change the appearance of the face to match a target lighting condition. This can be expressed through lighting parameters (e.g., an environment map, spherical harmonics, directional lighting, etc.) or through a reference image of another person. Our approach relights a portrait image to a lighting condition expressed through a low dynamic range (LDR) image representing the image on the monitor.

\noindent \textbf{Image based relighting.} Before the rise of deep learning, attempts were made at non deep learning image based relighting \cite{non_dl_method,non_dl_method1,non_dl_method2}. 
Shu \etal \cite{non_dl_method} introduced a face relighting approach that uses a mass-transport formulation for the transfer of illumination between images. Peers \etal \cite{non_dl_method1} demonstrated a method for relighting portrait images with flat lighting to match specific target environments, incorporating a reference subject database for approximation. Shih \etal \cite{non_dl_method2} adopted a multiscale technique to transfer local image statistics from reference portraits onto new ones, facilitating the matching of attributes like local contrast and overall lighting direction.
Recent advancements in deep learning have caused significant shifts to the landscape of portrait relighting \cite{DiFaReli,shadow_removal,blind_removal}. Research on ratio images for relighting \cite{ratio_paper1,ratio_paper2} was conducted, utilizing public datasets and employing methods based on ratio images. However, this is limited to synthetic data, resulting in a significant domain gap with real data. The widespread application of light stages~\cite{Light_stage} in gathering data has enabled numerous groundbreaking research endeavors\cite{Google,video_paper6,super_resolution_paper,harmonization}. However, these methods rely on capturing data with a light stage, which are expensive and inaccessible. Instead, our approach builds a personalized relighting algorithm using casually captured videos from the desk recording setup introduced in Sengupta \etal \cite{Roni}. In contrast to Sengupta \etal \cite{Roni}, we can collect data during daily computer usage by minimizing numerous constraints, eliminating the need for specific efforts in data collection.

\noindent \textbf{Relighting with explicit decomposition.} Creating a virtual relighting dataset through the utilization of synthetic human models to train their networks has been carried out in some studies \cite{sengupta2018sfsnet,explicit_paper5,explicit_paper8}. However, when using synthetic data to train a neural network, the large domain gap between synthetic and real data impacts the model's performance on real data. In contrast, others \cite{explicit_paper4,explicit_paper6,relitalk,relighting4d,nelf} generate relit images by using public datasets and employ methods based on 3D model rendering. Specifically, Hou \etal \cite{explicit_paper4} takes a more advanced approach by introducing explicit components, where rays originating from the face intersect with other parts of the facial geometry to create relit images.
Moreover, the extensive use of light stages~\cite{Light_stage} has enabled numerous innovative studies~\cite{explicit_paper1,explicit_paper2,explicit_paper3,explicit_paper7,explicit_paper9,diffusion_paper,texture_paper1,texture_paper2,super_resolution_paper,practical,learning} in this domain. Some researchers have incorporated explicit elements such as albedo, normals, specular maps, and diffuse maps into their methodologies \cite{explicit_paper1,explicit_paper2,explicit_paper3,diffusion_paper}. Others have taken a physics-based rendering approach \cite{explicit_paper7,explicit_paper9} to resolve these issues. Other papers\cite{texture_paper1,texture_paper2} aim to manipulate lighting conditions and generate images under different lighting scenarios using texture information. However, explicit decomposition methods require ground truth (GT) data for these intrinsic components to train, either from synthetic data or real light stage data. Obtaining GT data for `at-home' captures is impossible, thus we focus on image-based relighting.
In contrast, recent studies aim to streamline the capture process, using a mobile phone camera \cite{smartphone} or a sun stage \cite{sunstage} instead of a light stage. Nevertheless, due to their reliance on per-scene optimization, both of these papers lack the capability for real-time relighting and to generalize to unseen appearances of the individual and are only limited to the particular capture. Instead, our approach performs image-based relighting enabling real-time temporally consistent video relighting, at $\sim$45 fps.

\section{Method}
Our setup is similar to Gerstner \etal \cite{data_paper1} and Sengupta \etal \cite{Roni}, where a user's face is captured while illuminated by their monitor. By capturing multiple videos of the user's face along with the video on their monitor, we build our `at home' light stage dataset. We then use these data to train a personalized portrait relighting algorithm that can render the user's face under arbitrary lighting conditions. Specifically, given a portrait image $I_{\text{src}}$, corresponding source monitor lighting $L_{\text{src}}$, and target monitor lighting $L_{\text{trg}}$, our aim is to learn a function $G$ that relights $I_{\text{src}}$ under $L_{\text{trg}}$:
\begin{equation}\label{eq:model}
\hat{I}_{\text{trg}}, \hat{L}_{\text{src}} = G(I_{\text{src}}, L_{\text{src}}, L_{\text{trg}};\theta_G).
\end{equation}
Note that our formulation can be used for scenarios where the source lighting is unknown by simply replacing the input source lighting with the predicted source lighting, unlike previous approaches \cite{Roni}.

In the following sections, we outline our methodology for portrait video relighting using a monitor as a light stage. Sec. \ref{sec:data_pairing} outlines strategies for constructing training data pairs from casually captured videos that allow flexibility in facial expression, pose, and ambient lighting. In Sec. \ref{sec:network_architecture}, we introduce our relighting network architecture that disentangles lighting from intrinsic appearance using light-conditioned feature normalization, leading to high-quality relit images. In Sec. \ref{sec:temporal_consistency}, we propose additional techniques which enforce temporal consistency and eliminate flickering, also using LCFN. Finally, in Sec. \ref{sec:loss}, we discuss how to train our relighting network.

\subsection{Constructing training data pairs}\label{sec:data_pairing}
While past work \cite{Roni} imposed requirements of a neutral pose, expression, and dark room, we loosen these constraints to allow subjects in any conditions. The only constraint we maintain is that the room lighting shall not overpower the light emitted from the monitor. For example, if the capture occurs in front of a window with bright sunlight, the light from the monitor will have minimal impact on the face. As in Sengupta \etal \cite{Roni}, we aim to generate source and monitor image pairs $(I_{\text{src}},L_{\text{src}})$, as well as target image and target monitor pairs $(I_{\text{trg}},L_{\text{trg}})$, such that we can train our network to produce $\hat{I}_{\text{trg}}$ where  $I_{\text{trg}}$ is the pseudo ground truth. However, due to unrestricted subject movement during data collection, there is a lack of pixel-aligned data, making random pairs unsuitable. Previous work \cite{Roni} utilized segmentation for pairing.
However, we observed that segmentation is ineffective at finding pairs of images with the same pose and expression. Thus, we instead use facial keypoint detection \cite{facial_keypoint} to obtain source and target image pairs. 

\begin{figure*}[tb]
  \includegraphics[width=\linewidth]{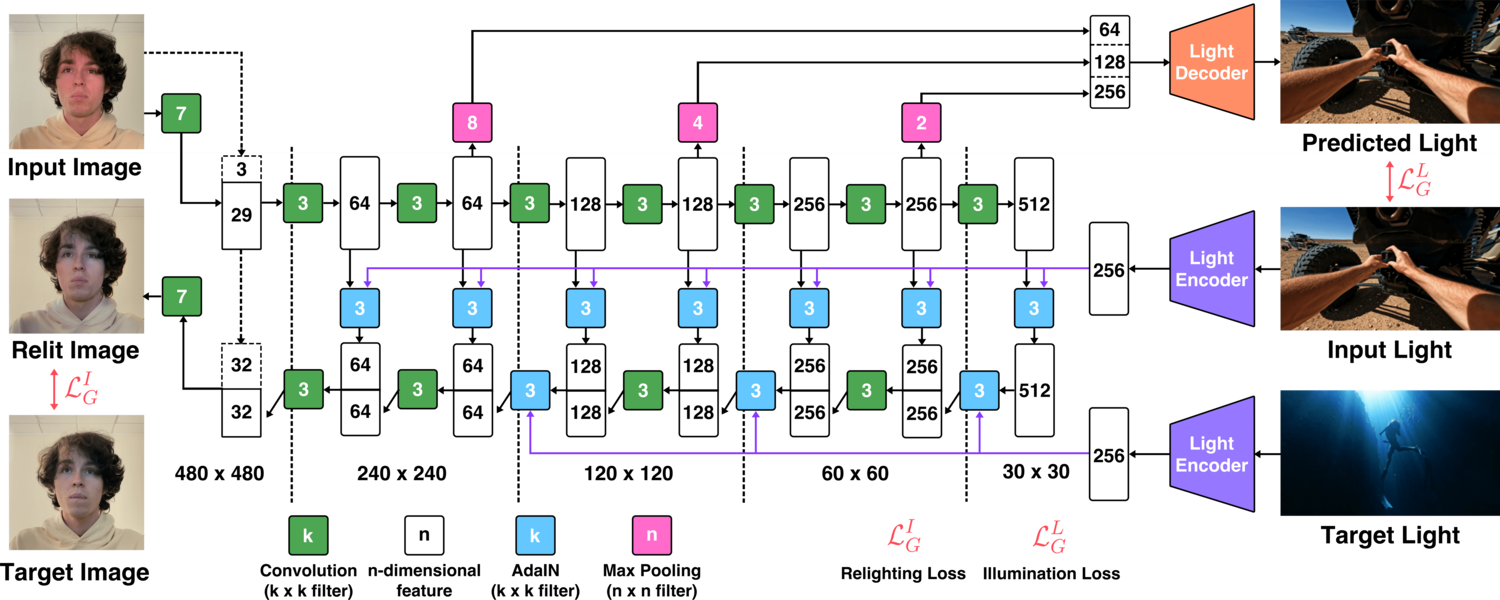}
  \caption{We first de-light the input image features extracted by the U-Net encoder using Adaptive Instance Normalization (AdaIN) guided by the lighting features extracted from the source lighting with a Light Encoder. We then pass these light-normalized encoder features to the decoder of the U-Net and apply another set of AdaIN guided by the features extracted from the target lighting with the Light Encoder. We additionally predict source lighting from the U-Net encoder using a Light Decoder.}
  \label{fig:model_architecture}
\end{figure*}

\subsection{Relighting network architecture}\label{sec:network_architecture}

Our network architecture, as illustrated in Fig. \ref{fig:model_architecture}, is built upon the well-established U-Net \cite{U-net}. This architecture is comprised of an encoder and a decoder with skip connections, which are commonly used in existing portrait relighting algorithms \cite{Google,Roni,ratio_paper1,ratio_paper2,super_resolution_paper}.
Our U-Net's encoder, similar to Sengupta \etal \cite{Roni} and Sun \etal \cite{Google}, processes the source portrait  $I_{\text{src}}$ by applying multiple convolutional layers of varying strides (1 or 2). This process progressively reduces spatial resolution while increasing the number of channels, yielding a latent feature space. The decoder performs the inverse function of the encoder by upsampling from the latent features and simultaneously skip-connecting to intermediate features from the encoder. These skip connections transport high-frequency shape and appearance information from the encoder to the decoder, ultimately resulting in the generation of a realistic relit image. However, they also carry source illumination features from the encoder to the decoder, leading to subpar relighting quality and temporal flickering. To address this issue, we introduce \textit{light-conditioned feature normalization} (LCFN) for the skip-connected features to better disentangle lighting features from intrinsic appearance features.

To disentangle lighting and intrinsic appearance components from the encoded features -- i.e. to de-light -- we first predict the source lighting from the encoder features. In contrast to Sun \etal \cite{Google}, which predicts the illumination $\hat{L}_{\text{src}}$ corresponding to the source image $I_{\text{src}}$ using the final encoded features, we take a different approach. We extract features at intermediate steps within the encoder, downsample them, and concatenate them using a confidence learning approach \cite{moni_prediction} to predict the illumination $\hat{L}_{\text{src}}$.

The LCFN module uses the lighting features generated by the lighting encoder to perform Adaptive Instance Normalization (AdaIN) \cite{Karras2019stylegan2} on the encoder features. We begin by using a multi-layer perceptron (MLP) to encode lighting features, transforming the lighting information of $L_{\text{src}}$ and $L_{\text{trg}}$ into a compact, low-dimensional representation (d = 256). We apply AdaIN to encoder features using the source lighting features, producing normalized features $f^l$ (for $l=1, \ldots, 7$). Through this normalization process, we induce de-lighting, effectively removing the lighting information present in the encoder features. Starting from the de-lit latent features ${f^7}$, we perform progressive bi-linear upsampling. At each upsampling step, we apply AdaIN to the concatenated feature, incorporating the target lighting features encoded by the lighting encoder. This construction using the LCFN module and source lighting prediction allows us to effectively remove source lighting features from the input and only propagate intrinsic appearance features from the encoder to the decoder. We then add target lighting features in the decoder. The LCFN module also contributes towards temporal consistency (see Sec. \ref{sec:temporal_consistency}). See Fig. \ref{fig:architecture_difference} for a comparison between our architecture and those of Sun \etal \cite{Google} and Sengupta \etal \cite{Roni}.

\subsection{Enforcing temporal consistency}\label{sec:temporal_consistency}
Temporal consistency is vital in making relit videos stable, realistic, and aesthetically pleasing. Previous single-image portrait relighting techniques \cite{Roni,Google} do not incorporate explicit temporal modeling, leading to undesirable flickering artifacts when applied to videos. Accuracy in single-image portrait relighting can often be uncorrelated to temporal flickering. Inconsistencies across frames are even more noticeable when the source lighting $L_{\text{src}}$ changes continuously.

When applied to skip-connected features, LCFN provides a natural defense against temporal flickering by removing source lighting features from the input image. However, it cannot ensure temporal consistency on its own. We notice two further problems: (1) when the source lighting gradually changes, LCFN often leaks small amounts of source lighting information to the decoder, leading to flickering; (2) when source lighting changes abruptly, undesirable fading effects can be observed.

To address this issue, we propose a \textit{skip-connected feature smoothing} technique that assumes neighboring frames share the same intrinsic appearance features, obtained after de-lighting input image features with LCFN. We apply a simple exponential smoothing of de-lit features generated by LCFN, denoted as $f^l$, using all the previous frames:
\begin{equation} 
    f_t^l \coloneqq \alpha \cdot f_t^l + (1-\alpha) \cdot f_{t-1}^l \quad(\text{for} \: l=1, \ldots, 7)
\end{equation}
with $\alpha = 0.7$. Note that exponential smoothing does not work without de-lit LCFN features, which removes time-varying source lighting.

We further notice that when the monitor light changes abruptly the relighting effect is delayed by a few frames, mainly due to the limited refresh rate of the monitor and frame rate of the camera. We thus propose doing a weighted average of source monitor lighting $L_{\text{src}}$ from a sequence of previous and current frames to achieve smoother and more natural results:
\begin{equation}
L^{t}_{\text{src avg}} = \frac{\sum_{i=0}^{N-1} \beta^i L^{t-i}_{\text{src}}}{\sum_{i=0}^{N-1} \beta^i},
\end{equation}
where $\beta=0.6$ and $N=3$.

\subsection{Training relighting network}\label{sec:loss}

Our model is trained through minimizing a weighted combination of three loss functions: generator loss, discriminator loss, and monitor loss. The first loss aims to minimize the discrepancies between the true target image $I_{\text{trg}}$ in our dataset and the predicted target relit image $\hat{I}_{\text{trg}}$, leading to accurately relit images. We adopted our generator loss (Eq.~\ref{eq:generator}) and our discriminator loss (Eq.~\ref{eq:discriminator}) from Sengupta \etal \cite{Roni}:
\begin{equation}\label{eq:generator}
\begin{split}
    \mathcal{L}_{G}^{I} &= \,  \lambda_{L1} \mathcal{L}_{L1}(I_{\text{trg}}, \hat{I}_{\text{trg}}) + \lambda_{P} \mathcal{L}_{P}(I_{\text{trg}}, \hat{I}_{\text{trg}}) \\
    & + \lambda_{C} \mathcal{L}_{C}(I_{\text{src}}, \hat{I}^{C}_{\text{src}}) + \lambda_{D} (D(\hat{I}_{\text{trg}};\theta_D)-1)^2,
\end{split}
\end{equation}
\begin{equation}\label{eq:discriminator}
\mathcal{L}_{D} = (D(I_{\text{trg}};\theta_D)-1)^2 + (D(\hat{I}_{\text{trg}};\theta_D))^2,
\end{equation}
where $\mathcal{L}_{L1}$ denotes L1 loss, $\mathcal{L}_{P}$ denotes perceptual loss \cite{perceptual_loss}, $\mathcal{L}_{C}$ denotes cycle consistency loss \cite{cycleGAN}, and $D$ is the discriminator \cite{patchGAN}. $\hat{I}_{\text{src}}^C$ and $\hat{L}_{\text{trg}}^C$ are the outputs from $G(\hat{I}_{\text{trg}}, L_{\text{trg}}, L_{\text{src}}, ; \theta_G)$

The monitor reconstruction loss focuses on minimizing the errors between the predicted source light $\hat{L}_{\text{src}}$ and the true source light $L_{\text{src}}$ and is expected to enforce improved disentanglement of lighting information from intrinsic appearance features.
\begin{equation}\label{eq:monitor_loss}
    \mathcal{L}_{G}^{M} = \lambda_{L1} \mathcal{L}_{L1}(L_{\text{src}}, \hat{L}_{\text{src}}) + \lambda_{P} \mathcal{L}_{P}(L_{\text{src}}, \hat{L}_{\text{src}})  + \lambda_{C} \mathcal{L}_{C}(L_{\text{trg}}, \hat{L}_{C}^{trg}).
\end{equation}

Finally, we minimize the image generator loss $\mathcal{L}_{G}^{I}$, the discriminator loss $\mathcal{L}_{D}$, and the illumination loss $\mathcal{L}_{G}^{M}$ together:
\begin{equation}\label{eq:total_loss}
\min_{G,D} \mathcal{L}_{G}^{I} + \mathcal{L}_{D} + \lambda_{G}^{M} \mathcal{L}_{G}^{M}.
\end{equation}

\begin{figure*}[tb]
  \includegraphics[width=\linewidth]{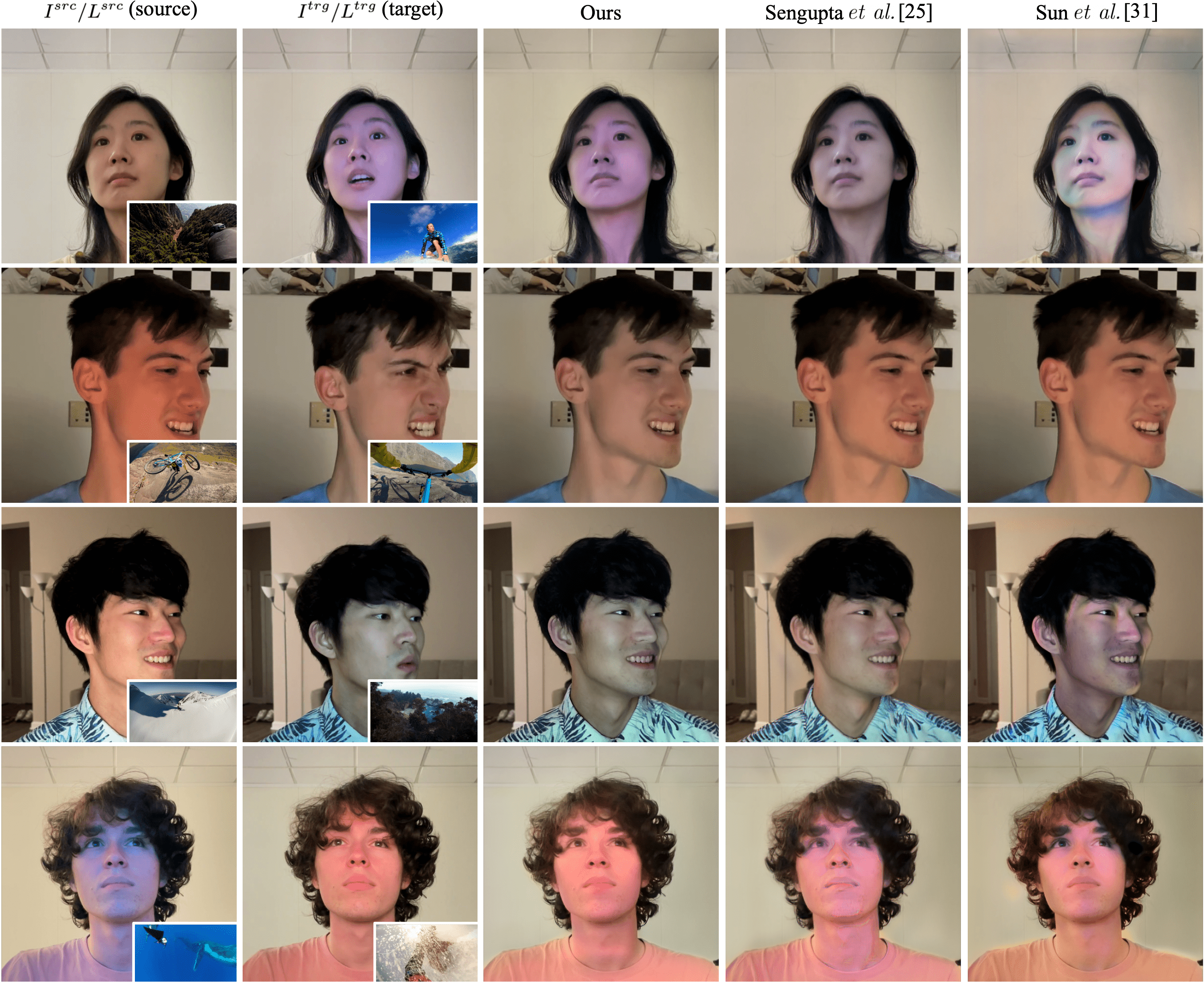}
  \caption{We perform a qualitative comparison with existing techniques \cite{Roni,Google} on the LSYD dataset. Source and target (images \& lighting) were unseen during training. All models are personalized, i.e. trained on images of that individual only. We (Col. 3) produce significantly better results compared to existing approaches (Cols 4 and 5).}
  \label{fig:visual_comparison}
\end{figure*}

\section{Experiments}
\label{sec:experiments}

In Sec. \ref{sec:Data}, we first discuss our data collection process, which was based on the approach used to compile the Light Stage at Your Desk (LSYD) dataset. In Sec. \ref{sec:metric}, we perform quantitative and qualitative comparisons with existing single-image portrait relighting algorithms and evaluate their temporal consistency. In Sec. \ref{sec:moni_pred}, we demonstrate predicting, at a low-level, what the user is looking at on the monitor screen. Finally, in Sec. \ref{sec:ablation}, we perform ablation studies, evaluating the network architecture's impact on relighting performance.

\begin{figure*}[tb]
  \includegraphics[width=\linewidth]{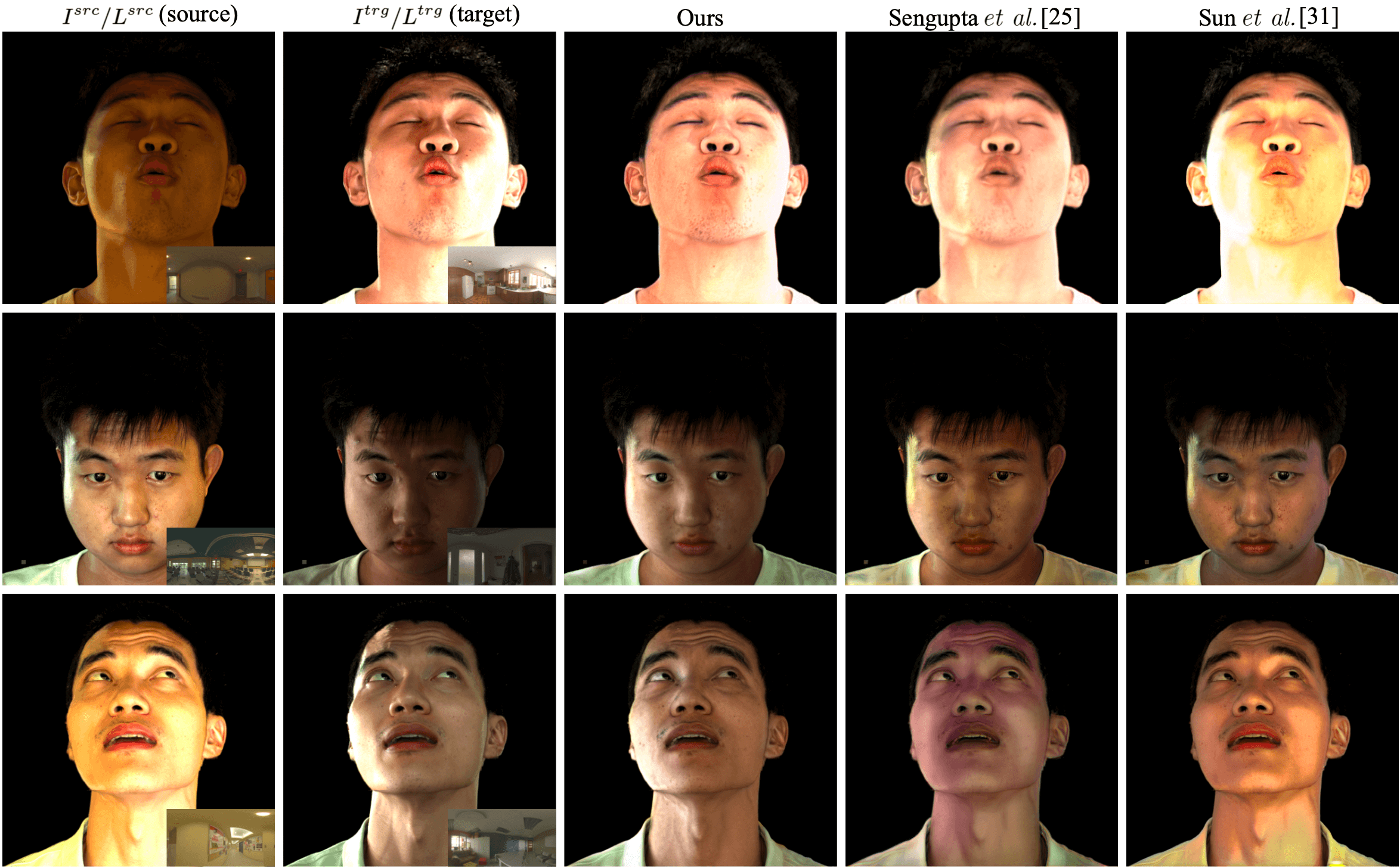}
  \caption{We perform a qualitative comparison on the OLAT dataset~\cite{video_paper6}. Our approach outperforms \cite{Google,Roni} and can render strong directional lighting and specular highlights without any explicit modeling of geometry and reflectance.}
  \label{fig:visual_comparison_olat}
\end{figure*}

\subsection{Data}\label{sec:Data}
We recorded data from 5 users of diverse ethnicities and genders to ensure a wide range of skin types. Each participant wore a variety of outfits, and we used 4 different ambient lighting conditions per person to mimic the conditions of real-life online meetings. We directed the participants to continuously change their facial expression and pose during the capture sessions. Each user's face was captured while watching 8 different videos, each 8 minutes long, on different days with varying appearances. We randomly hold out 1 video for testing and use the remaining 7 videos for training. We use this testing sequence only for qualitative evaluation, not for any quantitative metrics. This is because quantitative evaluation requires a pair of source and target images of the same person in the same pose but under different lighting conditions. This is difficult to obtain accurately for the aforementioned test video sequence since the participants naturally vary their pose and expression over the course of the video. Instead, we capture an additional test sequence, used only for numerical evaluation in which the participant is captured in 9 different pose-expression combinations, each with a distinct monitor light. For each pose, we can create $\binom{9}{2} = 36$ source and target pairs as input and pseudo ground-truth, resulting in a total of 324 test data pairs per user. 
Additionally, we compared methods using Dynamic OLAT Dataset~\cite{video_paper6} with environment lighting maps~\cite{laval_data1,laval_data2}. During this comparison, we converted HDR environment maps to LDR and utilized a 270$^{\circ}$ FoV as a monitor light to assess the relighting results. We want to note that all the test data is composed of unseen portraits and lightings for both LSYD and OLAT datasets.

\begin{table*}
  \caption{We train a personalized relighting model on 5 users from our LSYD dataset and 4 users from the OLAT dataset \cite{video_paper6}. We evaluate these models using source \& target portrait images, as well as lighting, that were not encountered during the training phase, and report average RMSE, LPIPS \cite{lpips}, and DISTS\cite{dists} scores both on LSYD and OLAT dataset. Our method can perform relighting without source lighting $L_{\text{src}}$, by simply using the predicted light source from our model as input lighting. Our method significantly outperforms  Sun \etal \cite{Google} and Sengupta \etal \cite{Roni}.}
  \label{tab:relit_comparison}
  \centering
  \resizebox{\linewidth}{!}{%
  \begin{tabular}{lccccccc}
    \toprule
    & Known source lighting & \multicolumn{3}{c}{\bfseries LSYD data} & \multicolumn{3}{c}{\bfseries OLAT data \cite{video_paper6}} \\
    \cmidrule(lr){3-5} \cmidrule(lr){6-8}
    & $\mathbf{L_{src}}$ &  \bfseries{LPIPS $\downarrow$} & \bfseries{DISTS $\downarrow$} & \bfseries{RMSE $\downarrow$} & \bfseries{LPIPS $\downarrow$} & \bfseries{DISTS $\downarrow$} & \bfseries{RMSE $\downarrow$}\\
    \midrule
    Sun \etal \cite{Google} w/ ${L_{\text{Sun}}}$  & \textendash & 0.1712 & 0.1629 & 8.5958  & 0.2273 & 0.1745 & 6.2692\\
    Sun \etal \cite{Google} w/ ${L_{\text{Ours}}}$& \textendash &  0.1029 & 0.1152 & 8.4476  & 0.2267 & 0.1569 & 6.1898\\
    Ours & \textendash &  0.0839 & 0.0953 & 8.3222 & 0.1812 & 0.1336 & 6.0931\\
    Sengupta \etal \cite{Roni} & \checkmark & 0.1018 & 0.1105 & 8.2826 & 0.2237 & 0.1675 & 6.1751\\
    Ours & \checkmark & \textbf{0.0832} & \textbf{0.0953} & \textbf{8.1939} & \textbf{0.1809} & \textbf{0.1334} & \textbf{5.9548}\\
  \bottomrule
  
\end{tabular}%
}
\end{table*}

\subsection{Comparison with existing approaches}\label{sec:metric}

We employ three error metrics to assess relighting performance: RMSE, LPIPS \cite{perceptual_loss}, and DISTS \cite{dists}. LPIPS and DISTS are more robust to slight differences in pose between the relit image and the pseudo ground truth and detect perceptual differences more effectively than RMSE.

\noindent \textbf{Portrait image relighting.} We compared our approach with existing portrait relighting neural architectures --- Sun \etal \cite{Google} and Sengupta \etal \cite{Roni} --- by training on our captured LSYD dataset using the same pre-processing for all three architectures (see Sec. \ref{sec:data_pairing}). Our training loss, given in Sec. \ref{sec:loss}, can handle misalignment in source-target pairs in training data, similar to the loss proposed in Sengupta \etal \cite{Roni} (we use an additional loss on source monitor lighting prediction). For Sun \etal \cite{Google}, we train both with their original loss function $L_{\text{Sun}}$ (which expects perfect source-target pose alignment obtained in OLAT data) and with our proposed loss function $L_{\text{ours}}$ to specifically handle misalignment in LSYD data. We train personalized models on 5 users from the LSYD dataset and on 4 users from the publicly available Dynamic OLAT Dataset~\cite{video_paper6} with 2361 indoor environment lighting maps~\cite{laval_data1,laval_data2}.

\setlength\intextsep{-2pt}
\begin{wrapfigure}[19]{r}{7cm}
    \centering
    \includegraphics[width=0.56\textwidth]{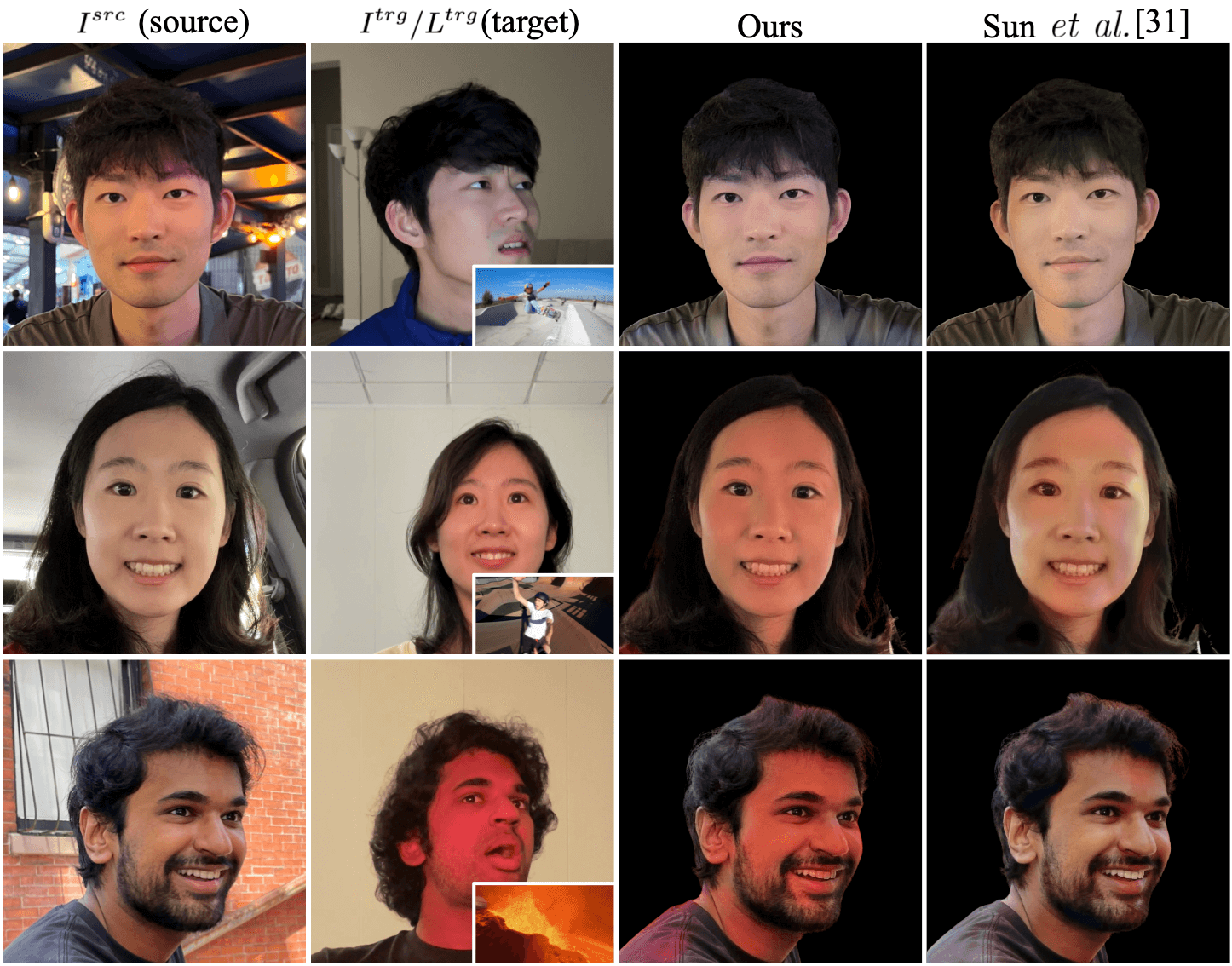}
    \caption{Our method can also relight portrait images captured with unknown light sources. We relight a source image (Col. 1) with target light shown in the inset in Col. 2. We add a reference of how the face appears under that target light.}
    \label{fig:in_the_wild}
\end{wrapfigure}

For our quantitative evaluation, we test our model on 1620 test images across 5 users with unseen appearance and lighting conditions on the LSYD dataset and on 7172 test images from the Dynamic OLAT dataset. We present the result in Tab. \ref{tab:relit_comparison}. We observe that our proposed approach outperforms Sengupta \etal \cite{Roni} and Sun \etal \cite{Google} by 22.3\% and 23.6\% respectively on the LSYD dataset and by 23.5\% and 25.6\% on the OLAT dataset, when comparing LPIPS score. Our qualitative comparison, as presented in Fig. \ref{fig:visual_comparison} and Fig. \ref{fig:visual_comparison_olat}, shows that our model performs superior relighting in terms of color, quality, and consistency, and can render strong directional lighting and specular highlights without any explicit modeling.

Note that Sun \etal \cite{Google} does not require the source lighting $L_{\text{src}}$ during test time. Our proposed approach can also perform relighting without prior knowledge of source lighting $L_{\text{src}}$ by simply predicting $\hat{L}_{\text{src}}$ and using it for light-conditioned feature normalization. We show that even in the absence of $L_{\text{src}}$, our method outperforms Sun \etal \cite{Google} by 22.6\% on the LSYD data and by 25.5\% on the OLAT data, in terms of LPIPS. 

In Fig. \ref{fig:in_the_wild}, we demonstrate that our approach can relight any portrait image captured `in-the-wild' without requiring the knowledge of source lighting, and outperforms Sun \etal \cite{Google}. Our approach first predicts a proxy source monitor lighting, imagining the portrait image to be captured under a monitor lighting, and uses this predicted source lighting in the LCFN module to `de-lit' the input image and relight it with a target lighting.

\setlength\intextsep{-4pt}
\begin{wrapfigure}[14]{r}{7cm}
    \centering
    \includegraphics[width=0.56\textwidth]{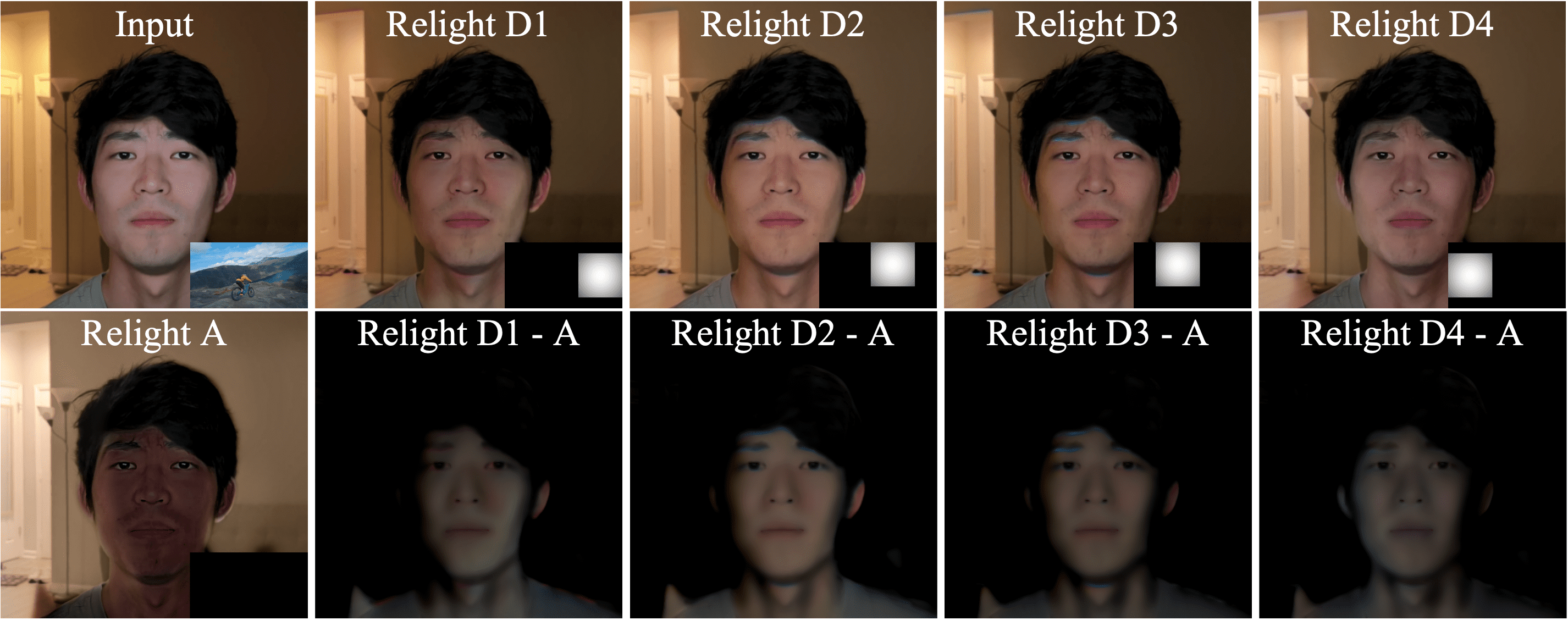}
    \caption{We relight the input image with directional lights (1st row) and without any target light (`Relight A'). We then subtract the `Relight A' from directional images (row-2 col-2:5). Our method learns to render the effects of directional lighting and decouple ambient and dominant frontal lighting (from monitor).}
    \label{fig:directional_light}
\end{wrapfigure}

In Fig. \ref{fig:directional_light} 1st-row, we show the results of relighting an input image with directional lights by moving specific bright areas on the monitor screen. Our method learns to render directional lighting effects and cast shadows as needed. To better illustrate the effect of moving lighting, we first relight the input image with no light reflected from the monitor (row-2 col-1), which produces a relit image under ambient room lighting only. Subtracting this `de-lit' image from the relit images under directional lighting (row-2 col 2-5) highlights the ability of our method to decouple ambient room illumination from dominant frontal lighting. However, due to the constraint that our lighting is limited to the illumination emitted from the monitor, we may not accurately depict extreme lighting effects. To this end, we show the strong directional light effects using the OLAT dataset in Fig. \ref{fig:visual_comparison_olat}.

\setlength\intextsep{-10pt}
\begin{wraptable}[17]{r}{6.6cm}
\scriptsize
\centering
\caption{We evaluate temporal consistency by relighting a test video captured with varying source lighting with the same target lighting and calculating RMSE between adjacent frames. We then report the average RMSE across all adjacent frames and compute an error rate to indicate the percent of adjacent frames with RMSE higher than a threshold.}
\resizebox{\linewidth}{!}{%
\begin{tabular}{lccccc}
    \toprule
    & \bfseries{\shortstack{RMSE $\downarrow$}} & \multicolumn{3}{c}{\bfseries Error Rate (\%)}\\
    \cmidrule(lr){3-5}
    Threshold & & $>$0.2 & $>$0.3 & $>$0.4\\
    \midrule
    Sun \etal \cite{Google} &  5.86 & 13.53 & 2.61 & 1.06\\
    \cmidrule{1-1}
    Sengupta \etal \cite{Roni} &  6.37 & 21.83 & 5.40 & 1.75\\
    +$L_{\text{src\_avg}}$ &  5.76 & 13.31 & 2.34 & 0.98\\
    \cmidrule{1-1}
    Ours (base)  &  6.01 & 16.22 & 3.61 & 1.08\\
    +$L_{\text{src\_avg}}$  & 5.73 & 13.09 & 2.31& 0.83\\
    +LCFN  & 5.68 & 13.04 & 2.28 & 0.71\\
    +$L_{\text{src\_avg}}$+LCFN  & \textbf{5.55} & \textbf{12.89} & \textbf{2.22} & \textbf{0.65}\\
    \bottomrule
  \end{tabular}%
  }
\label{tab:consistency_comparison}
\end{wraptable}

\noindent \textbf{Portrait video relighting.} Next, we evaluate the temporal consistency of each portrait video relighting algorithm. For each user in the LSYD data, we relit the held-out test video with 50 different target lighting conditions, creating 50 relit videos. We then computed the RMSE between adjacent frames in relit videos as a measure of temporal consistency. Since the pose is almost identical between adjacent frames, lower RMSE error indicates temporally consistent relighting. We then report the average temporal RMSE across all such adjacent frames. In practice, however, a significant fraction of adjacent frame pairs have extremely similar lighting between the two frames, making their relit frames naturally consistent anyway. Only in a small percentage of adjacent frames does the source lighting significantly change, leading to obvious flickering in the relit video if temporal consistency is not maintained. Thus, in addition to average temporal RMSE, we also compute the error rate for three different thresholds:  0.2 (low), 0.3 (medium), and 0.4 (high), which indicate the percentage of adjacent frames where RMSE error is more than the threshold.

\setlength\intextsep{-4pt}
\begin{wrapfigure}[23]{r}{7cm}
    \centering
    \includegraphics[width=0.56\textwidth]{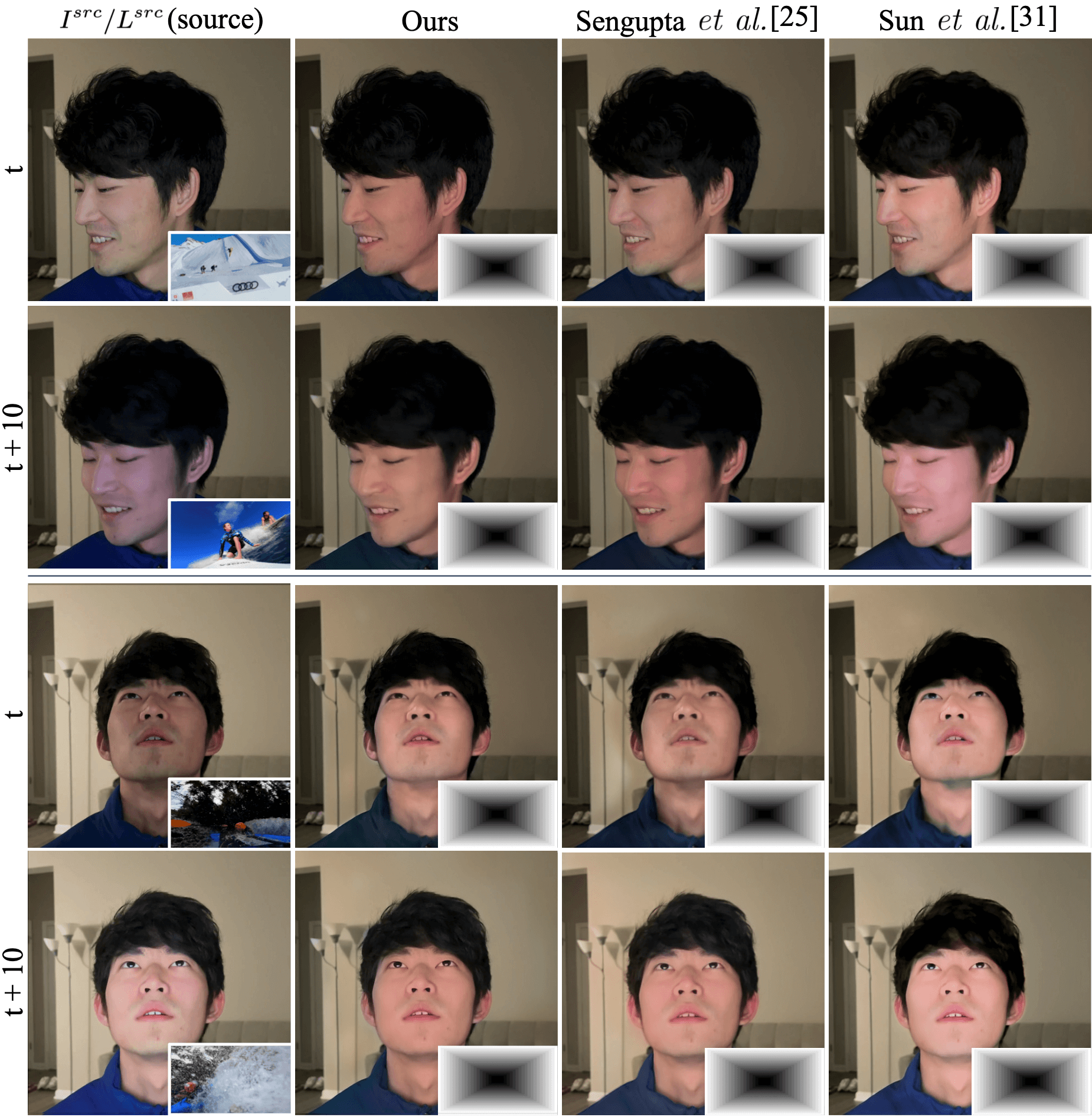}
    \caption{We show temporal consistency between adjacent frames separated by 0.33s by relighting a test video captured with varying source lighting with the same target lighting. Note that \cite{Google,Roni} both exhibit abrupt changes in lighting between frames $t$ and $t+10$, while our approach produces a more stable result.}
    \label{fig:temporal}
\end{wrapfigure}

In Tab. \ref{tab:consistency_comparison} and Fig. \ref{fig:temporal}, we compare our approach with and without skip-connected feature smoothing to past works \cite{Google,Roni}. Note that this temporal smoothing of the skip-connected features can only be applied in our framework since we de-light encoder features from the source lighting with LCFN. Both for our approach and for Sengupta \etal \cite{Roni}, we can further apply smoothing of input source lighting to handle abrupt changes. We observe that our method produces the most temporally consistent relighting while also being the most accurate (see Tab. \ref{tab:relit_comparison}). We further note that both skip-connected feature smoothing and smoothing of source lighting improve temporal consistency.

\subsection{Monitor prediction}\label{sec:moni_pred}

\setlength\intextsep{-4pt}

\begin{wrapfigure}[10]{r}{7cm}
    \centering
    \includegraphics[width=0.56\textwidth]{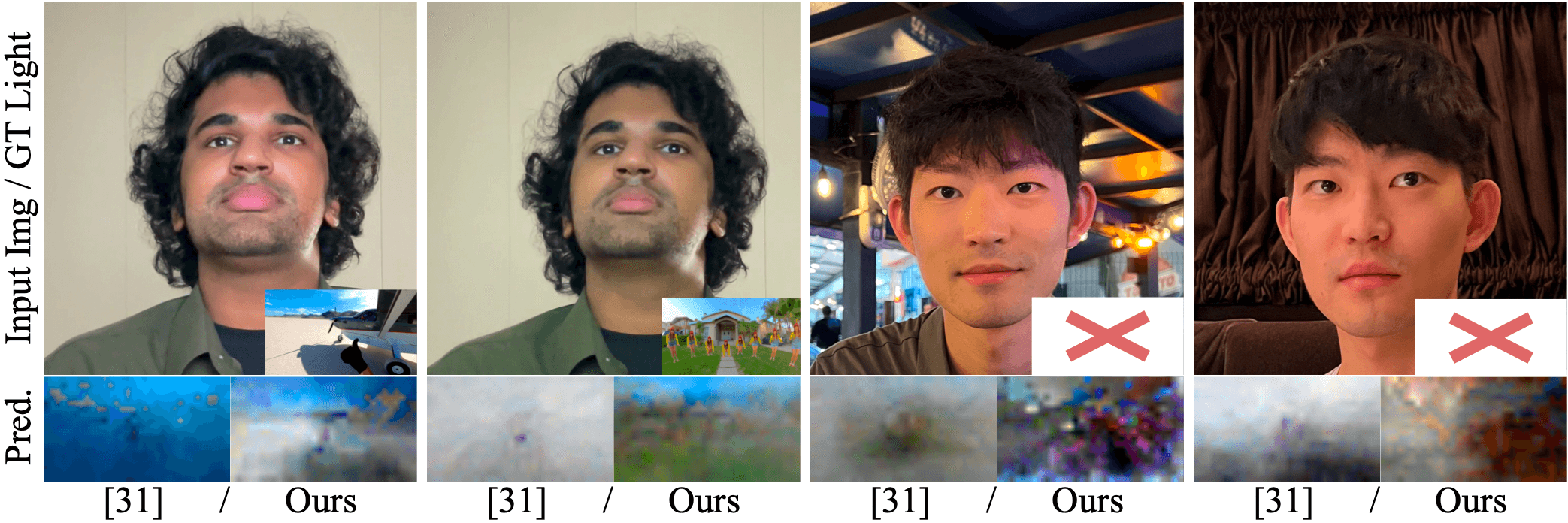}
    \caption{Monitor prediction comparison on LSYD data (Col. 1 and 2) and ``in the wild" setting (Col. 3 and 4). The second row represents the monitor prediction by \cite{Google} and Ours.}
    \label{fig:moni_pred}
\end{wrapfigure}

Our network can also be used to predict the source monitor lighting of any input image, as shown in Fig. \ref{fig:moni_pred}. For the LSYD dataset, we do have images of the source monitor for every input image, so we can calculate monitor prediction accuracy numerically. We observe that our method slightly outperforms \cite{Google} with a Mean Absolute Error of 0.15 vs 0.17, also qualitatively producing more meaningful visualizations. For portrait images, we do not have ground-truth source lighting, but visualizations show meaningful predictions.

The ability to predict monitor lighting from images captured by our monitor has many implications. This technique can be used to detect deep fake avatars during live video calls by purposefully projecting specific images via screen sharing and observing if we can detect the same image from the webcam feed of other attendees in the call. If there is a mismatch between the projected monitor image and the predicted one, this likely indicates a deep fake avatar in the video call.

\subsection{Ablation studies}\label{sec:ablation}

\setlength\intextsep{-10pt}
\begin{wraptable}[9]{r}{6.6cm}
\scriptsize
\centering
\caption{Both LCFN and source monitor prediction $\mathbf{L_{src}}$ improve relighting performance by effectively disentangling source lighting information from intrinsic appearance features.}
\resizebox{\linewidth}{!}{%
\begin{tabular}{ccccc}
        \toprule
        $\mathbf{L_{src}}$ & \bfseries{de-lighting} & \bfseries{RMSE $\downarrow$} & \bfseries{LPIPS $\downarrow$} & \bfseries{DISTS $\downarrow$} \\
        \midrule
        \textendash & \textendash & 8.4230 & 0.0964 & 0.1071 \\
        \checkmark & \textendash & 8.2596 & 0.0915 & 0.1013 \\
        \textendash & \checkmark & 8.1907 & 0.0904 & 0.0966 \\
        \checkmark & \checkmark & \textbf{8.0746} & \textbf{0.0853} & \textbf{0.0963} \\
        \bottomrule
    \end{tabular}%
    }
\label{tab:relit_ablations}
\end{wraptable}
We report the removal of various components from our relighting network in Tab. \ref{tab:relit_ablations}, specifically LCFN and source monitor lighting prediction using intermediate encoder features. We observe that both improve final relighting performance, which shows their effectiveness in disentangling source lighting information from intrinsic appearance features.

\section{Conclusion}

We propose a personalized video relighting algorithm that leverages casually captured LSYD data to generate real-time high-quality temporally consistent relit videos under any pose, expression, and lighting conditions. We present a novel network architecture that can perform high-quality relighting on both the LSYD and OLAT datasets, rendering challenging lighting conditions like directional lights, shadows, and specularities. We achieve this without using any 3D information or performing explicit decomposition, simply by achieving better image-based relighting enabled by our proposed neural architecture with LCFN module. Our method enables better lighting quality during live video calls and in portrait images, and produces better harmonization with virtual backgrounds.

\noindent \textbf{Limitation.} Since we utilize only the front-facing monitor light as the source lighting, we do not account for scenarios where the light source is located on the sides (90$^\circ$) or behind (180$^\circ$). However, it does not necessarily mean that our model cannot learn directional lighting effects, as demonstrated by training on the OLAT dataset which contains full 360$^\circ$ lighting, see Fig. \ref{fig:visual_comparison_olat}.

\noindent \textbf{Ethical considerations.} While our primary goal is to allow people to improve their facial appearance with virtual relighting, we note that it is also a form of image manipulation and can be used for malicious purposes \cite{secure}. Furthermore, we emphasize that while currently we can only predict the user's direct line-of-sight monitor lighting in low resolution, the potential for high-resolution monitor prediction in the future could raise significant privacy concerns.

\par\vfill\par

%
%
\bibliographystyle{splncs04}
\bibliography{main}

\end{document}